\documentclass[runningheads]{llncs}

\usepackage[T1]{fontenc}

\usepackage{graphicx}
\usepackage{booktabs}
\usepackage{amssymb}
\usepackage{multirow}
\usepackage{bbding}
\begin{document}

\title{SAUF-Net: Structure--Appearance Representation Learning with Uncertainty Feedback for Semi-Supervised Medical Image Segmentation}
\titlerunning{SAUF-Net for Semi-Supervised Medical Image Segmentation}

\author{Qin Lu \and
Zheyang Jing \and
Yujie Yang \and
Jianwang Li \and
Chen Yi \and
Shaofeng Jiang\Envelope}
\authorrunning{Q. Lu et al.}
%
\institute{Nanchang Hangkong University, Nanchang, China\\
\email{jsphone@163.com}}
\maketitle              

\begin{abstract}
Semi-supervised learning has shown great potential for reducing annotation costs in medical image segmentation. However, most existing methods mainly exploit unlabeled data through prediction-level consistency, while the reliability of internal
feature representations is often overlooked. In
medical images, target-related structural cues are
easily entangled with unstable appearance
variations, which may lead to unreliable pseudo
labels and error accumulation during training. To
address these issues, we propose SAUF-Net, a
Structure--Appearance Representation Learning with
Uncertainty Feedback Network for semi-supervised
medical image segmentation. SAUF-Net uses the
Structure--Appearance Decomposition Module (SADM) to
separate bottleneck features into structural and
appearance representations. The Disentangled
Guidance Module (DGM) injects these representations into the decoding
process to enhance structure-aware segmentation. Meanwhile, the Auxiliary Decoder produces branch-specific predictions for reliability
estimation and a fused prediction for
appearance-swapped consistency. Furthermore, we introduce an
Appearance-Swapped Consistency branch to encourage
structural representations to remain stable under
appearance variations. We also introduce a
reliability-map-guided dual-head discriminator with
a Validity Head and an Uncertainty Head to provide
feature-level uncertainty feedback. Extensive
experiments on ISIC-2016 and Kvasir-SEG demonstrate
that SAUF-Net outperforms state-of-the-art semi-supervised methods, especially under low-label
settings.

\keywords{Semi-supervised Learning \and Medical
Image Segmentation \and Structure--Appearance
Learning \and Feature Disentanglement \and
Uncertainty Feedback}
\end{abstract}
\section{Introduction}
{\let\thefootnote\relax\footnotetext{Q.Lu and Z.Jing—Co-first authors.}}
In recent years, deep learning models, including
Convolutional Neural Networks (CNNs)
\cite{ronneberger2015u} and Transformers
\cite{chen2021transunet}, have achieved remarkable
progress in medical image segmentation. These
models show strong potential in automatically
delineating anatomical structures and lesions.
However, most fully supervised segmentation methods
still rely on large-scale, high-quality pixel-level
annotations \cite{cao2022swin,10183842}. In
clinical practice, obtaining such annotations is
expensive and time-consuming because it requires
expert knowledge and careful manual delineation.
This annotation burden limits the practical
application of fully supervised methods, especially
in label-scarce scenarios.

Semi-supervised learning (SSL) provides a promising
solution by jointly using a small set of labeled
images and a large set of unlabeled images
\cite{chen2021semi,tarvainen2017mean,yao2022enhancing}. Existing SSL methods mainly exploit unlabeled
data through consistency regularization, pseudo
label generation, and uncertainty-aware sample
selection. Multi-stream or multi-task consistency
methods encourage stable predictions under
different perturbations
\cite{Ouali_2020_CVPR,10.1007/978-3-030-59710-8_54,HE2025103626}, distribution-alignment methods
reduce the gap between labeled and unlabeled data \cite{CHARTSIAS2019101535,Bai_2023_CVPR,Zhang_2025_CVPR}, and reliability-aware methods refine pseudo-labels by estimating prediction uncertainty \cite{10.1007/978-3-030-32245-8_67,NEURIPS2023_c28ef844,NEURIPS2023_1f7e6d5c,11150469}. Although these strategies improve the use of
unlabeled data, most of them regularize training
mainly at the prediction level, such as enforcing
output consistency or selecting pseudo-labels
according to confidence. Such prediction-level
constraints can make outputs more stable, but they
do not explicitly ensure that the internal feature
representations are reliable. This limitation is particularly important for
medical images, where target-related structural
cues, such as lesion shape, object location, and
boundary continuity, are often entangled with
unstable appearance cues, such as background
texture, color variation, imaging artifacts, and
local noise. When labeled data are scarce, the
model may produce confident pseudo-labels while
still relying on unstable appearance cues rather
than segmentation-relevant structural information.
Once these biased predictions are used for
training, the errors may be reinforced on unlabeled
data, leading to overconfident mistakes and
progressive pseudo-label error accumulation.

To address these issues, we propose the Structure--Appearance Representation Learning with Uncertainty Feedback Network (SAUF-Net) for semi-supervi-sed
medical image segmentation. Instead of relying only
on prediction-level regularization, SAUF-Net
improves unlabeled-data learning by modeling
structure--appearance representations and
introducing reliability-guided feature-level
feedback. The main contributions of this work are
summarized as follows:
(1) We propose a Structure--Appearance Decomposition Module (SADM) to separate target-related structural cues from unstable appearance variations. The decomposed representations are injected into the Disentangled Guidance Module (DGM) for decoding, while the Auxiliary Decoder provides auxiliary predictions for reliability estimation.
(2) We introduce an Appearance-Swapped Consistency (ASC) branch to enhance structural robustness. By pairing structural representations with shuffled appearance representations, ASC encourages the model to learn structure-centered features that remain stable under appearance changes.  
(3) We develop a reliability-guided uncertainty feedback branch to reduce pseudo-label error accumulation on unlabeled data. A dual-head discriminator estimates feature validity and feature-level uncertainty under reliability guidance, improving the quality of unlabeled representations. 
Experiments on ISIC-2016 and Kvasir-SEG validate the effectiveness of SAUF-Net under low-label settings.
\section{Methods}
\subsection{Overview}
\begin{figure}
    \centering
    \vspace{-4mm}
    \includegraphics[width=1\linewidth]{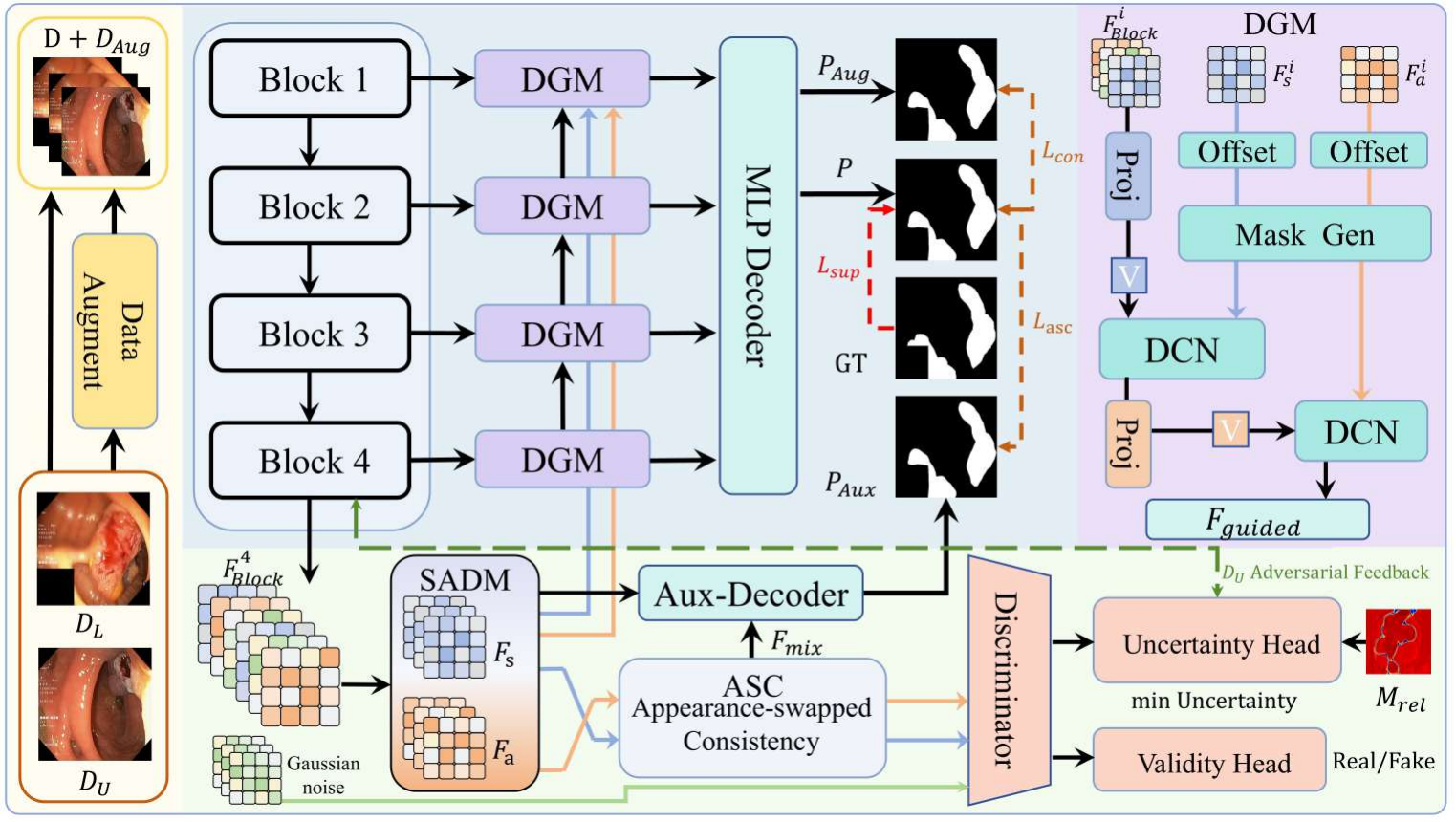}
    \caption{Illustration of the proposed SAUF-Net. Blue and orange solid lines denote the data flows of the structural representation $F_s$ and the appearance representation $F_a$, respectively. In the Appearance-Swapped Consistency (ASC) branch, the blue and orange lines indicate the original non-swapped $F_s$ and $F_a$ streams, while the output after appearance swapping is denoted as the mixed feature $F_{mix}$.}
    \label{fig1}
\vspace{-4mm}
\end{figure}
In the semi-supervised setting, the training data
consist of a labeled set $\mathcal{D}_L=\{(x_i^l,y_i^l)\}_{i=1}^{N_l}$ and an unlabeled set $\mathcal{D}_U=\{x_i^u\}_{i=1}^{N_u}$, where $N_l \ll N_u$. We denote the complete training set as $\mathcal{D}=\mathcal{D}_L\cup\mathcal{D}_U$. During training, an augmented set $\mathcal{D}_{Aug}$ is constructed from $\mathcal{D}$ using color or wavelet transformation. Both $\mathcal{D}$ and $\mathcal{D}_{Aug}$ are fed into the shared segmentation network to obtain
the original prediction $P$ and the augmented
prediction $P_{Aug}$. 

As shown in Fig.~\ref{fig1}, SAUF-Net adopts a
SegFormer-B4-based architecture. The encoder
extracts multi-scale features $\{F_{\mathrm{Block}}^i\}_{i=1}^4$ and the bottleneck
feature $F_{\mathrm{Block}}^4$ is decomposed by the
Structure--Appearance Decomposition Module (SADM)
into a structural representation $F_s$ and an
appearance representation $F_a$. These
representations are used by the Disentangled
Guidance Module (DGM) to guide decoding and
generate the final prediction $P$. Meanwhile, the Auxiliary Decoder (Aux-Decoder) produces the branch-specific predictions $P_s$ and
$P_a$ for reliability estimation, together with a
fused prediction $P_{Aux}$ for ASC. For unlabeled data, $P$ is used to generate pseudo
labels, and the reliability map $M_{\mathrm{rel}}$ is
estimated from $P$, $P_s$, and $P_a$ to guide the
learning of the augmented prediction $P_{Aug}$. In
addition, the Appearance-Swapped Consistency (ASC)
branch constructs mixed features $F_{mix}$ by
pairing structural representations with shuffled
appearance representations, while the uncertainty
feedback branch employs a dual-head discriminator,
consisting of a Validity Head and an Uncertainty
Head, to estimate feature validity and spatial uncertainty
under the guidance of $M_{\mathrm{rel}}$. 

\subsection{Structure--Appearance Representation Learning}

Given an input image $x$, an augmented image
$x_{Aug}$ is constructed by applying either color
transformation or wavelet-based transformation. The
original image $x$ and the augmented image
$x_{Aug}$ are fed into the same segmentation
network with shared parameters, producing the
prediction $P$ and the augmented prediction
$P_{Aug}$, respectively.

The shared SegFormer-B4 encoder extracts multi-scale features $\{F_{\mathrm{Block}}^i\}_{i=1}^4$, where $F_{\mathrm{Block}}^{i}$ denotes the feature from the
$i$-th encoder stage. SADM contains
two parallel projection branches with independent
parameters. Each branch consists of two $3\times3$
Conv-BN-ReLU blocks followed by one $1\times1$ Conv-BN-ReLU block. The two branches generate the
structural representation $F_s$ and the appearance
representation $F_a$ from the deepest encoder
feature $F_{\mathrm{Block}}^4$, respectively. The
structural representation $F_s$ is designed to
capture target-related cues, such as object shape,
spatial location, region continuity, and boundary
structure, which are directly related to
segmentation. In contrast, the appearance
representation $F_a$ models complementary
appearance information, such as background texture,
intensity variation, color distribution, and
imaging noise. This decomposition encourages the
network to rely more on stable structural cues
while reducing the influence of unstable appearance
variations. 

To provide explicit structure--appearance
guidance for multi-scale feature reconstruction, we
introduce the Disentangled Guidance Module (DGM) to
inject $F_s$ and $F_a$ into the decoding process.
Before entering the $i$-th DGM, $F_s$ and $F_a$ are resized to the same spatial resolution as the current decoding feature by bilinear upsampling and convolutional refinement, and the resized representations are denoted as $F_s^i$ and $F_a^i$, respectively. In the deepest decoding
stage, DGM takes $F_{\mathrm{Block}}^{4}$ and the resized
structural and appearance representations as inputs.
In the remaining stages, the current encoder
feature is concatenated with the upsampled guided
feature from the previous deeper stage, while $F_s^i$ and $F_a^i$ serve as guidance features.
For the $i$-th DGM, let $X^i$ denote the input
decoding feature, which is first projected into a
unified feature representation $V^i=\mathrm{Proj}(X^i)$. Then, DGM refines $V^i$ through two serial
deformable guidance steps. First, the resized
structural representation $F_s^i$ is used to
generate the offset and modulation mask:

\begin{equation}
\Delta_s^i = \mathrm{Conv}_{off}^{s}(F_s^i), \qquad M_s^i = \sigma(\mathrm{Conv}_{mask}^{s}(F_s^i)).
\end{equation}
The projected feature $V^i$ is sampled by deformable
convolution with the offset $\Delta_s^i$ and mask $M_s^i$, producing the structure-guided feature $\widetilde{F}_s^i$. This step enables the decoder to adaptively focus on structure-related regions. Next, the
structurally guided feature is further refined by
the resized appearance representation $F_a^i$.
Similarly, $F_a^i$ generates another offset and
modulation mask:
\begin{equation}
\Delta_a^i = \mathrm{Conv}_{off}^{a}(F_a^i), \qquad M_a^i = \sigma(\mathrm{Conv}_{mask}^{a}(F_a^i)).
\end{equation}
The final guided feature of this stage is obtained
by:
\begin{equation}
F_{guided}^{i}=
\mathrm{DCN}(\mathrm{Proj}(\widetilde{F}_s^i);\Delta_a^i,M_a^i).
\end{equation}
Through this serial structure-to-appearance
guidance, DGM adaptively refines the decoding
feature using the decomposed representations.

The guided features from different stages are aggregated by the MLP decoder to generate the segmentation prediction $P$. Meanwhile, the Aux-Decoder contains structural,
appearance, and fusion branches. The first two
branches generate $P_s=g_s(F_s)$ and
$P_a=g_a(F_a)$, respectively, which are used for
reliability estimation. The fusion branch takes the
channel-wise concatenation of $F_s$ and $F_a$ and
produces
$P_{Aux}=H([F_s,F_a])$.
Each branch progressively upsamples its input using
bilinear interpolation and Conv-BN-ReLU blocks,
followed by a $1\times1$ prediction layer.

\subsection{Appearance-Swapped Consistency (ASC)}

Although SADM separates structural and appearance
representations, $F_s$ may still contain
appearance-dependent information. To
preserve segmentation structure under appearance
replacement, we introduce Appearance-Swapped Consistency (ASC).

Given two samples $A$ and $B$ in a mini-batch, ASC
keeps $F_s^A$ and replaces $F_a^A$ with the
shuffled appearance representation $F_a^B$. The
mixed representation and its fused prediction are
defined as:
\begin{equation}
F_{mix}
=
[F_s^A,F_a^B],
\qquad
P_{Aux}^{mix}
=
H(F_{mix}),
\end{equation}
Here, $P_{Aux}^{mix}$ denotes the output
generated from the appearance-swapped pair
$(F_s^A,F_a^B)$, whereas $P_{Aux}$ in the
previous subsection is generated from the original
pair $(F_s^A,F_a^A)$. Both predictions are produced
by the same fusion branch $H$. For simplicity, 
Fig.~\ref{fig1} uses $P_{Aux}$ to denote
this shared auxiliary prediction branch. ASC
therefore encourages the fused prediction to
preserve the structure provided by $F_s^A$ despite
changes in its paired appearance representation. 

\subsection{Reliability-Guided Uncertainty Feedback}

Pseudo-labels may be unreliable in ambiguous or
low-contrast regions. We therefore introduce a
reliability-guided uncertainty feedback branch to
regularize unlabeled representations.
Given the main prediction $P$ and the
branch-specific predictions $P_s$ and $P_a$, the
reliability map is defined as:
\begin{equation}
M_{\mathrm{rel}}
=
(1-|P-P_s|)\cdot|P_s-P_a|.
\end{equation}
The first term measures the agreement between the
main and structural predictions, whereas the
second measures the response contrast between the
structural and appearance branches. Larger values
therefore indicate more reliable regions.

As shown in Fig.~\ref{fig1}, a shared dual-head
discriminator $D$ is applied separately to $F_s$
and $F_a$. Its Validity Head $D_v$ distinguishes
encoder-derived features from synthetic features
produced by feature generators from Gaussian
latent vectors. Features extracted from labeled
and unlabeled images are treated as real, whereas
the generated features are treated as fake.

The Uncertainty Head $D_u$ predicts a spatial
uncertainty map. For labeled data, the structural
and appearance targets are computed from the
prediction errors of $P_s^l$ and $P_a^l$ against
$y^l$ and $1-y^l$, respectively. For unlabeled
data, the detached complementary reliability map
$1-M_{\mathrm{rel}}^u$ is resized to the feature
resolution and used as the uncertainty target.
The resulting discriminator feedback encourages
augmented unlabeled features to be both valid and
low-uncertainty.

\subsection{Training Objective}

For labeled images, given an image $x^l$ and its ground-truth mask $y^l$, the network produces the main prediction $P^l$,
the branch-specific predictions $P_s^l$ and
$P_a^l$, and the fused auxiliary prediction
$P_{Aux}^l$. The basic segmentation loss $\mathcal{L}_{seg}$ is implemented by combining Dice loss and BCE loss. The main prediction $P^l$ and the structural auxiliary prediction $P_s^l$ are supervised by the foreground mask, while the appearance auxiliary prediction $P_a^l$ is supervised by the complementary background mask. The supervised loss is defined as:
\begin{equation}
\mathcal{L}_{sup} =
\mathcal{L}_{seg}(P^l,y^l)
+\lambda_{aux}
\left[
\mathcal{L}_{seg}(P_s^l,y^l)
+\mathcal{L}_{seg}(P_a^l,1-y^l)
+\mathcal{L}_{seg}(P_{Aux}^l,y^l)
\right].
\end{equation}

For each unlabeled sample, the original image $x^u$
is fed into the network to obtain the main
prediction $P^u$, the auxiliary predictions $P_s^u$
and $P_a^u$, and the reliability map
$M_{\mathrm{rel}}^u$. All predictions denote
sigmoid-normalized probabilities. The binary
pseudo-label is defined as
$\hat{y}_j^u=\mathbb{I}(P_j^u>0.5)$, where $j$
indexes pixels. The confidence mask is defined as
$M_{\mathrm{conf},j}^u=\mathbb{I}
(\max(P_j^u,1-P_j^u)>\tau)$, where $\tau=0.95$.
The final pixel-wise weight is
$W_j^u=M_{\mathrm{conf},j}^u\cdot
\mathrm{sg}(M_{\mathrm{rel},j}^u)$, where
$\mathrm{sg}(\cdot)$ denotes the stop-gradient
operation. The pseudo-label and pixel-wise weight
are computed without gradient propagation and are
used to supervise the augmented prediction
$P_{Aug}^u$.

The augmented unlabeled image $x_{Aug}^u$ is fed into the same network to obtain the augmented prediction $P_{Aug}^u$, which is supervised by the pseudo-label:
\begin{equation}
\mathcal{L}_{con}=
\frac{\sum_{j}
\left(W_j^u\mathcal{L}_{bce}(P_{Aug,j}^{u},\hat{y}_{j}^{u})
\right)}{
\sum_{j} W_j^u+\epsilon
}.
\end{equation}

For appearance-swapped supervision, the structural representation is kept unchanged, while the appearance representation is replaced by a shuffled appearance feature. The mixed representation is processed by the
fusion branch to produce
$P_{Aux}^{mix,u}$, which is
supervised by the pseudo-label associated with its
structural representation:

\begin{equation}
\mathcal{L}_{asc}=
\frac{\sum_j W_j^u
\mathcal{L}_{bce}
(P_{Aux,j}^{mix,u},
\hat{y}_{j}^{u})}
{\sum_j W_j^u+\epsilon}.
\end{equation}

For uncertainty feedback, the shared discriminator
is optimized using adversarial classification and
uncertainty regression. Encoder-derived features
are treated as real, while generator-synthesized
features are treated as fake. The Uncertainty Head
is supervised by the labeled prediction errors and
the resized complementary reliability map for
unlabeled data. The feature generators are
adversarially trained to produce features classified
as real.

When optimizing the segmentation network, the
discriminator is fixed. Let $\widetilde{F}_s^u$ and
$\widetilde{F}_a^u$ denote the augmented unlabeled
features. The uncertainty feedback loss is:
\begin{equation}
\mathcal{L}_{uf}
=
\frac{1}{2}\sum_{k\in\{s,a\}}
\left[
\mathcal{L}_{adv}
(D_v(\widetilde{F}_k^u),1)
+
\mathcal{L}_{mse}
(D_u(\widetilde{F}_k^u),0)
\right].
\end{equation}
The final objective for optimizing the segmentation network is:
\begin{equation}
\mathcal{L}_{total}=
\mathcal{L}_{sup}+
\lambda_{con}(t)\mathcal{L}_{con}+
\lambda_{asc}\mathcal{L}_{asc}+
\lambda_{uf}(t)\mathcal{L}_{uf}.
\end{equation}

Here, $\lambda_{con}(t)$ and $\lambda_{uf}(t)$ are ramp-up weights. $\lambda_{asc}$ controls the contribution of the ASC loss. During training, the discriminator and feature generator are optimized alternately with the segmentation network. During inference, only the segmentation network is retained.

\section{Experiments}
\subsection{Datasets and Evaluation Metrics}
\textbf{Datasets.} We evaluate our method on two public medical image segmentation datasets: (1) ISIC-2016 \cite{DBLP:journals/corr/GutmanCCHMMH16}: This dataset contains dermoscopy images for skin cancer diagnosis with corresponding pixel-level annotations, comprising 900 training images and 379 testing images. (2) Kvasir-SEG \cite{10.1007/978-3-030-37734-2_37}: This dataset includes 1,000 polyp images with ground-truth labels. We randomly allocate 80\% of the images for training and the remaining 20\% for testing.

\textbf{Evaluation Metrics.} To comprehensively evaluate the performance of all models, we employ the Dice Similarity Coefficient (DSC), Intersection over Union (IoU), and pixel-level Accuracy (Acc).

\subsection{Implementation Details}
We adopt SegFormer-B4 pre-trained on ImageNet as the encoder backbone. All input images are resized to $512\times512$, and the batch size is set to 2. The network is optimized using AdamW with a weight decay of 0.0001. The training process follows a two-stage strategy. First, the network is trained on labeled data for 100 epochs with a learning rate of 0.0001. Then, unlabeled data are introduced for semi-supervised training for 50 epochs with a learning rate of 0.00001. Augmented images in $\mathcal{D}_{Aug}$ are generated using either color or wavelet transformation. The color transformation includes random color jittering and grayscale conversion, while the wavelet-based transformation perturbs the image in the frequency domain. The auxiliary loss weight and appearance-swapped supervision weight are set to $\lambda_{aux}=0.4$ and $\lambda_{asc}=0.05$, respectively. The consistency weight is set to $\lambda_{con}(t)=0.1\cdot\mathrm{Rampup}(t)$, and the uncertainty feedback weight is defined as $\lambda_{uf}(t)=\alpha\lambda_{con}(t)$, with
$\alpha=0.10$. The pseudo-label confidence threshold is set to 0.95. The feature generators and shared discriminator are optimized using Adam with a learning rate of $2\times10^{-4}$. All experiments are implemented in PyTorch on a single NVIDIA GeForce RTX 4070 Ti SUPER GPU.

\begin{table}[!t]
\centering
\caption{Comparison of segmentation performance on ISIC-2016 and Kvasir-SEG datasets. The best results are highlighted in bold.}
\label{tab1}
\vspace{-1.5mm}
\scriptsize
\renewcommand{\arraystretch}{0.85}
\setlength{\tabcolsep}{2.2pt}
\resizebox{0.99\linewidth}{!}{
\begin{tabular}{|l|c|ccc|ccc|}
\hline
Methods & Ratio & \multicolumn{3}{c|}{ISIC-2016} & \multicolumn{3}{c|}{Kvasir-SEG} \\
\cline{3-8}
& & Dice (\%)$\uparrow$ & IoU (\%)$\uparrow$ & Acc (\%)$\uparrow$ & Dice (\%)$\uparrow$ & IoU (\%)$\uparrow$ & Acc (\%)$\uparrow$ \\
\hline
SegFormer B4 \cite{NEURIPS2021_64f1f27b} & 100\% & 92.88 & 87.31 & 95.93 & 91.79 & 86.86 & 97.48 \\
\hline
SegFormer B4 \cite{NEURIPS2021_64f1f27b} & 10\% & 87.15 & 79.54 & 93.51 & 78.83 & 71.07 & 94.52 \\
MT \cite{tarvainen2017mean} & 10\% & 88.01 & 81.42 & 94.03 & 83.45 & 76.32 & 94.97 \\
SASSNet \cite{10.1007/978-3-030-59710-8_54} & 10\% & 88.42 & 81.46 & 93.78 & 82.78 & 75.21 & 94.78 \\
ST++ \cite{Yang_2022_CVPR} & 10\% & 88.17 & 82.46 & 94.98 & 86.23 & 80.41 & 95.31 \\
CCT \cite{Ouali_2020_CVPR} & 10\% & 88.22 & 81.34 & 93.78 & 83.74 & 76.45 & 94.88 \\
CPS \cite{chen2021semi} & 10\% & 88.89 & 82.12 & 94.28 & 82.81 & 76.34 & 94.91 \\
SLCNet \cite{10.1007/978-3-031-16452-1_14} & 10\% & 88.23 & 82.09 & 94.82 & 83.43 & 75.64 & 94.79 \\
DMT \cite{feng2022dmt} & 10\% & 88.03 & 82.34 & 94.93 & 84.98 & 77.23 & 94.89 \\
BCP \cite{Bai_2023_CVPR} & 10\% & 89.02 & 82.93 & 95.09 & 85.79 & 78.43 & 95.09 \\
ARCO \cite{NEURIPS2023_1f7e6d5c} & 10\% & 89.52 & 84.26 & 95.45 & 87.96 & 81.52 & 96.45 \\
DAW \cite{NEURIPS2023_c28ef844} & 10\% & 89.67 & 84.31 & 95.52 & 88.24 & 81.63 & 96.54 \\
AdaptFRCNet \cite{HE2025103626} & 10\% & \underline{91.37} & \underline{85.30} & \underline{95.70} & \underline{89.35} & \underline{83.36} & \underline{96.71} \\
\textbf{Ours} & \textbf{5\%} & \textbf{90.91} & \textbf{84.76} & \textbf{95.56} & \textbf{89.71} & \textbf{83.48} & \textbf{96.95} \\
\textbf{Ours} & \textbf{10\%} & \textbf{91.88} & \textbf{85.88} & \textbf{96.11} & \textbf{90.53} & \textbf{84.94} & \textbf{96.93} \\
\hline
SegFormer B4 \cite{NEURIPS2021_64f1f27b} & 20\% & 88.65 & 81.58 & 94.32 & 82.16 & 74.69 & 95.27 \\
MT \cite{tarvainen2017mean} & 20\% & 89.83 & 83.15 & 95.43 & 84.45 & 77.43 & 95.47 \\
SASSNet \cite{10.1007/978-3-030-59710-8_54} & 20\% & 89.94 & 83.67 & 95.19 & 83.97 & 77.21 & 95.56 \\
ST++ \cite{Yang_2022_CVPR} & 20\% & 90.21 & 85.23 & 95.59 & 87.94 & 82.17 & 96.36 \\
CCT \cite{Ouali_2020_CVPR} & 20\% & 89.59 & 82.99 & 95.42 & 84.89 & 77.94 & 95.78 \\
CPS \cite{chen2021semi} & 20\% & 89.03 & 84.67 & 95.11 & 85.83 & 78.86 & 95.88 \\
SLCNet \cite{10.1007/978-3-031-16452-1_14} & 20\% & 89.34 & 84.54 & 95.34 & 85.42 & 78.35 & 95.74 \\
DMT \cite{feng2022dmt} & 20\% & 89.21 & 84.32 & 95.56 & 86.47 & 79.32 & 95.83 \\
BCP \cite{Bai_2023_CVPR} & 20\% & 89.98 & 85.01 & 95.32 & 87.43 & 80.45 & 96.23 \\
ARCO \cite{NEURIPS2023_1f7e6d5c} & 20\% & 92.43 & 86.21 & 96.28 & 90.19 & 84.85 & 96.89 \\
DAW \cite{NEURIPS2023_c28ef844} & 20\% & 92.31 & 86.13 & 96.27 & 90.54 & 84.96 & 96.93 \\
AdaptFRCNet \cite{HE2025103626} & 20\% & \underline{92.53} & \underline{86.83} & \underline{96.29} & \underline{90.71} & \underline{85.23} & \underline{97.32} \\
\textbf{Ours} & \textbf{20\%} & \textbf{92.64} & \textbf{86.89} & \textbf{96.47} & \textbf{91.98} & \textbf{86.69} & \textbf{97.59} \\
\hline
\end{tabular}
}
\vspace{-2mm}
\end{table}

\subsection{Comparison with State-of-the-Art Methods}

We compare SAUF-Net with recent SOTA methods on both datasets. Our experiments follow the same configuration as AdaptFRCNet~\cite{HE2025103626}. As shown in Table~\ref{tab1}, at a 10\% labeled ratio, SAUF-Net achieves a Dice of 91.88\% and an IoU of 85.88\% on ISIC-2016, outperforming AdaptFRCNet by 0.51\% and 0.58\%, respectively. On Kvasir-SEG, SAUF-Net obtains a Dice of 90.53\% and an IoU of 84.94\%, improving over AdaptFRCNet by 1.18\% and 1.58\%. Moreover, with only 5\% labeled data, SAUF-Net achieves 90.91\% Dice on ISIC-2016 and 89.71\% Dice on Kvasir-SEG, which is competitive with or better than many methods trained with 10\% labeled data. These results indicate that structure--appearance representation learning and reliability-guided uncertainty feedback can improve the utilization of unlabeled images. When the labeled ratio increases to 20\%, SAUF-Net still outperforms AdaptFRCNet on both datasets, improving Dice by 0.11\% on ISIC-2016 and 1.27\% on Kvasir-SEG. The improvements in per-image Dice over AdaptFRCNet were statistically significant across both datasets and all evaluated labeling ratios (all \(p<0.01\); two-sided Wilcoxon signed-rank tests).
\begin{figure}
    \centering
    \includegraphics[width=1\linewidth]{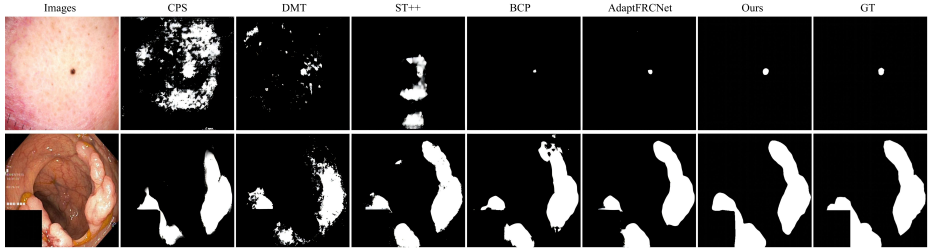}
    \caption{The segmentation results of skin-lesion segmentation (first row) and polyp (second row) tasks using 10\% labeled data.}
    \label{fig2}
\vspace{-4mm}
\end{figure}

Fig.~\ref{fig2} visualizes challenging segmentation cases. Compared with other methods, SAUF-Net better preserves target structures and boundary details, especially for small lesions, blurry boundaries, and complex background textures. This suggests that the proposed structure--appearance learning and uncertainty feedback help reduce unreliable predictions in ambiguous regions.
\subsection{Ablation Study}
\begin{table}[!t]
\centering
\caption{Ablation studies and hyper-parameter analysis on Kvasir-SEG with 20\% labeled data.}
\label{tab:ablation_all}
\scriptsize
\renewcommand{\arraystretch}{1.02}
\setlength{\tabcolsep}{3pt}

\textbf{(a) Ablation of SAUF-Net components.}
\vspace{0.5mm}

\resizebox{0.88\linewidth}{!}{
\begin{tabular}{@{}ccccc ccc@{}}
\toprule
\multicolumn{5}{c}{\textsc{Components}} & \multicolumn{3}{c}{\textsc{Metrics}} \\
\cmidrule(r){1-5} \cmidrule(l){6-8}
\textsc{Baseline} & \textsc{SADM} & \textsc{ASC} & $M_{\mathrm{rel}}$ & \textsc{UF} & Dice (\%)$\uparrow$ & IoU (\%)$\uparrow$ & Acc (\%)$\uparrow$ \\
\midrule
$\surd$ &         &         &         &         & 82.16 & 74.69 & 95.27 \\
$\surd$ & $\surd$ &         &         &         & 89.91 & 84.77 & 96.39 \\
$\surd$ & $\surd$ &         & $\surd$ &         & 91.16 & 85.44 & 96.74 \\
$\surd$ & $\surd$ & $\surd$ & $\surd$ &         & 91.38 & 85.86 & 96.99 \\
$\surd$ & $\surd$ &         & $\surd$ & $\surd$ & 90.54 & 85.01 & 97.29 \\
$\surd$ & $\surd$ & $\surd$ & $\surd$ & $\surd$ & \textbf{91.98} & \textbf{86.69} & \textbf{97.59} \\
\bottomrule
\end{tabular}
}

\vspace{2mm}

\begin{minipage}[t]{0.31\linewidth}
\centering
\textbf{(b) Effect of ASC weight $\lambda_{asc}$.}
\vspace{0.5mm}

\begin{tabular}{@{}ccc@{}}
\toprule
$\lambda_{asc}$ & Dice (\%) & IoU (\%) \\
\midrule
0.03 & 90.72 & 85.18 \\
0.05 & \textbf{91.98} & \textbf{86.69} \\
0.10 & 90.94 & 85.74 \\
\bottomrule
\end{tabular}
\end{minipage}%
\hfill
\begin{minipage}[t]{0.31\linewidth}
\centering
\textbf{(c) Effect of feedback coefficient $\alpha$.}
\vspace{0.5mm}

\begin{tabular}{@{}ccc@{}}
\toprule
$\alpha$ & Dice (\%) & IoU (\%) \\
\midrule
0.07 & 91.16 & 85.73 \\
0.10 & \textbf{91.98} & \textbf{86.69} \\
0.13 & 90.82 & 85.25 \\
\bottomrule
\end{tabular}
\end{minipage}%
\hfill
\begin{minipage}[t]{0.31\linewidth}
\centering
\textbf{(d) Effect of confidence threshold $\tau$.}
\vspace{0.5mm}

\begin{tabular}{@{}ccc@{}}
\toprule
$\tau$ & Dice (\%) & IoU (\%) \\
\midrule
0.90 & 91.76 & 86.36 \\
0.95 & \textbf{91.98} & \textbf{86.69} \\
0.98 & 91.81 & 86.40 \\
\bottomrule
\end{tabular}
\end{minipage}
\vspace{-4mm}
\end{table}

\textbf{Effectiveness of structure--appearance decomposition.}
Table~\ref{tab:ablation_all}(a) reports the component ablation results on Kvasir-SEG with 20\% labeled data. Compared with the baseline SegFormer-B4, SADM improves the Dice from 82.16\% to 89.91\% and the IoU from 74.69\% to 84.77\%. This demonstrates that separating structural cues from appearance variations provides more discriminative representations for segmentation. By explicitly modeling structural information, the model can better focus on target shape, spatial location, and boundary continuity rather than being dominated by unstable texture or intensity changes.

\textbf{Effectiveness of reliability-guided learning.}
For variants without $M_{\mathrm{rel}}$, we use the conventional maximum-probability confidence strategy with $\tau=0.95$. Thus, these variants replace the proposed reliability map with standard confidence rather than removing pseudo-label filtering. Introducing $M_{\mathrm{rel}}$ improves the Dice and IoU from 89.91\% and 84.77\% to 91.16\% and 85.44\%, respectively. This result indicates that $M_{\mathrm{rel}}$ provides more informative reliability estimates than conventional prediction confidence.

\textbf{Effectiveness of ASC and uncertainty feedback.}
Adding ASC improves the Dice from 91.16\% to 91.38\% and the IoU from 85.44\% to 85.86\%, showing that appearance-swapped supervision enhances the robustness of structural representations. The uncertainty feedback branch alone improves Acc to 97.29\%, but its Dice and IoU are lower than those of the full model, suggesting that feature-level uncertainty feedback is more effective when the structural representation has been regularized by ASC. With all components enabled, SAUF-Net achieves the best performance, with 97.59\% Acc, 91.98\% Dice, and 86.69\% IoU. These results confirm the complementarity of SADM, ASC, $M_{\mathrm{rel}}$, and uncertainty feedback.

\textbf{Hyper-parameter sensitivity.}
Table~\ref{tab:ablation_all}(b)-(d) analyzes the effects of $\lambda_{asc}$, the uncertainty feedback coefficient $\alpha$, and the pseudo-label confidence threshold $\tau$. The best ASC performance is obtained at $\lambda_{asc}=0.05$, indicating that an appropriate ASC weight can effectively regularize structural representations. For uncertainty feedback, $\alpha=0.10$ achieves the best Dice and IoU, showing that a stronger but controlled feedback weight improves feature-level refinement. For pseudo-label filtering, $\tau=0.95$ achieves the best performance, suggesting that overly loose or strict confidence filtering may weaken unlabeled-data learning. These results show that the selected hyper-parameters provide a good balance between supervision strength and training stability.

\section{Conclusion}
We propose SAUF-Net, a structure--appearance representation learning framework with reliability-guided uncertainty feedback for semi-supervised medical image segmentation. SAUF-Net decomposes bottleneck features through SADM and injects them into multi-scale decoding with DGM. ASC preserves segmentation structure under appearance variations, while uncertainty feedback reduces the influence of unreliable pseudo-labels. Experiments on ISIC-2016 and Kvasir-SEG show competitive performance under low-label settings. However, the current evaluation is limited to 2D binary lesion segmentation and does not assess cross-dataset generalization. Future work will extend SAUF-Net to 3D and multi-class tasks and evaluate its robustness across datasets and imaging domains.

\begin{credits}
\subsubsection{\ackname} This work was supported by the National Natural Science Foundation of China under Grant 62261039.

\subsubsection{\discintname}
The authors have no competing interests to declare that are relevant to the content of this article.
\end{credits}

\bibliographystyle{splncs04}
\bibliography{reference}

@inproceedings{ronneberger2015u,
  title={U-net: Convolutional networks for biomedical image segmentation},
  author={Ronneberger, Olaf and Fischer, Philipp and Brox, Thomas},
  booktitle={International Conference on Medical image computing and computer-assisted intervention},
  pages={234--241},
  year={2015},
  organization={Springer}
}

@article{chen2021transunet,
  title={Transunet: Transformers make strong encoders for medical image segmentation},
  author={Chen, Jieneng and Lu, Yongyi and Yu, Qihang and Luo, Xiangde and Adeli, Ehsan and Wang, Yan and Lu, Le and Yuille, Alan L and Zhou, Yuyin},
  journal={arXiv preprint arXiv:2102.04306},
  year={2021}
}

@inproceedings{cao2022swin,
  title={Swin-unet: Unet-like pure transformer for medical image segmentation},
  author={Cao, Hu and Wang, Yueyue and Chen, Joy and Jiang, Dongsheng and Zhang, Xiaopeng and Tian, Qi and Wang, Manning},
  booktitle={European conference on computer vision},
  pages={205--218},
  year={2022},
  organization={Springer}
}

@ARTICLE{10183842,
  author={Zhou, Hong-Yu and Guo, Jiansen and Zhang, Yinghao and Han, Xiaoguang and Yu, Lequan and Wang, Liansheng and Yu, Yizhou},
  journal={IEEE Transactions on Image Processing}, 
  title={nnFormer: Volumetric Medical Image Segmentation via a 3D Transformer}, 
  year={2023},
  volume={32},
  number={},
  pages={4036-4045}}

@inproceedings{chen2021semi,
  title={Semi-supervised semantic segmentation with cross pseudo supervision},
  author={Chen, Xiaokang and Yuan, Yuhui and Zeng, Gang and Wang, Jingdong},
  booktitle={Proceedings of the IEEE/CVF conference on computer vision and pattern recognition},
  pages={2613--2622},
  year={2021}
}

@article{tarvainen2017mean,
  title={Mean teachers are better role models: Weight-averaged consistency targets improve semi-supervised deep learning results},
  author={Tarvainen, Antti and Valpola, Harri},
  journal={Advances in neural information processing systems},
  volume={30},
  year={2017}
}

@inproceedings{yao2022enhancing,
  title={Enhancing pseudo label quality for semi-supervised domain-generalized medical image segmentation},
  author={Yao, Huifeng and Hu, Xiaowei and Li, Xiaomeng},
  booktitle={Proceedings of the AAAI conference on artificial intelligence},
  volume={36},
  pages={3099--3107},
  year={2022}
}

@article{CHARTSIAS2019101535,
title = {Disentangled representation learning in cardiac image analysis},
journal = {Medical Image Analysis},
volume = {58},
pages = {101535},
year = {2019},
issn = {1361-8415},
author = {Agisilaos Chartsias and Thomas Joyce and Giorgos Papanastasiou and Scott Semple and Michelle Williams and David E. Newby and Rohan Dharmakumar and Sotirios A. Tsaftaris}
}

@InProceedings{Bai_2023_CVPR,
    author    = {Bai, Yunhao and Chen, Duowen and Li, Qingli and Shen, Wei and Wang, Yan},
    title     = {Bidirectional Copy-Paste for Semi-Supervised Medical Image Segmentation},
    booktitle = {Proceedings of the IEEE/CVF Conference on Computer Vision and Pattern Recognition (CVPR)},
    month     = {June},
    year      = {2023},
    pages     = {11514-11524}
}

@InProceedings{Ouali_2020_CVPR,
author = {Ouali, Yassine and Hudelot, Celine and Tami, Myriam},
title = {Semi-Supervised Semantic Segmentation With Cross-Consistency Training},
booktitle = {Proceedings of the IEEE/CVF Conference on Computer Vision and Pattern Recognition (CVPR)},
month = {June},
year = {2020}
}

@inproceedings{10.1007/978-3-031-16452-1_14,
author = {Liu, Jinhua and Desrosiers, Christian and Zhou, Yuanfeng},
title = {Semi-supervised Medical Image Segmentation Using Cross-Model Pseudo-Supervision with Shape Awareness and Local Context Constraints},
year = {2022},
isbn = {978-3-031-16451-4},
publisher = {Springer-Verlag},
address = {Berlin, Heidelberg},
booktitle = {Medical Image Computing and Computer Assisted Intervention – MICCAI 2022: 25th International Conference, Singapore, September 18–22, 2022, Proceedings, Part VIII},
pages = {140–150},
numpages = {11},
location = {Singapore, Singapore}
}

@InProceedings{10.1007/978-3-030-59710-8_54,
author="Li, Shuailin
and Zhang, Chuyu
and He, Xuming",
editor="Martel, Anne L.
and Abolmaesumi, Purang
and Stoyanov, Danail
and Mateus, Diana
and Zuluaga, Maria A.
and Zhou, S. Kevin
and Racoceanu, Daniel
and Joskowicz, Leo",
title="Shape-Aware Semi-supervised 3D Semantic Segmentation for Medical Images",
booktitle="Medical Image Computing and Computer Assisted Intervention -- MICCAI 2020",
year="2020",
publisher="Springer International Publishing",
address="Cham",
pages="552--561",
isbn="978-3-030-59710-8"
}

@InProceedings{Yang_2022_CVPR,
    author    = {Yang, Lihe and Zhuo, Wei and Qi, Lei and Shi, Yinghuan and Gao, Yang},
    title     = {ST++: Make Self-Training Work Better for Semi-Supervised Semantic Segmentation},
    booktitle = {Proceedings of the IEEE/CVF Conference on Computer Vision and Pattern Recognition (CVPR)},
    month     = {June},
    year      = {2022},
    pages     = {4268-4277}
}

@article{HE2025103626,
title = {AdaptFRCNet: Semi-supervised adaptation of pre-trained model with frequency and region consistency for medical image segmentation},
journal = {Medical Image Analysis},
volume = {103},
pages = {103626},
year = {2025},
issn = {1361-8415},
author = {Along He and Yanlin Wu and Zhihong Wang and Tao Li and Huazhu Fu}
}

@InProceedings{Zhang_2025_CVPR,
    author    = {Zhang, Zheng and Yin, Guanchun and Zhang, Bo and Liu, Wu and Zhou, Xiuzhuang and Wang, Wendong},
    title     = {A Semantic Knowledge Complementarity based Decoupling Framework for Semi-supervised Class-imbalanced Medical Image Segmentation},
    booktitle = {Proceedings of the IEEE/CVF Conference on Computer Vision and Pattern Recognition (CVPR)},
    month     = {June},
    year      = {2025},
    pages     = {25940-25949}
}

@InProceedings{10.1007/978-3-030-32245-8_67,
author="Yu, Lequan
and Wang, Shujun
and Li, Xiaomeng
and Fu, Chi-Wing
and Heng, Pheng-Ann",
editor="Shen, Dinggang
and Liu, Tianming
and Peters, Terry M.
and Staib, Lawrence H.
and Essert, Caroline
and Zhou, Sean
and Yap, Pew-Thian
and Khan, Ali",
title="Uncertainty-Aware Self-ensembling Model for Semi-supervised 3D Left Atrium Segmentation",
booktitle="Medical Image Computing and Computer Assisted Intervention -- MICCAI 2019",
year="2019",
publisher="Springer International Publishing",
address="Cham",
pages="605--613",
isbn="978-3-030-32245-8"
}

@inproceedings{NEURIPS2023_c28ef844,
 author = {Sun, Rui and Mai, Huayu and Zhang, Tianzhu and Wu, Feng},
 booktitle = {Advances in Neural Information Processing Systems},
 editor = {A. Oh and T. Naumann and A. Globerson and K. Saenko and M. Hardt and S. Levine},
 pages = {61792--61805},
 publisher = {Curran Associates, Inc.},
 title = {DAW: Exploring the Better Weighting Function for Semi-supervised Semantic Segmentation},
 volume = {36},
 year = {2023}
}

@inproceedings{NEURIPS2023_1f7e6d5c,
 author = {You, Chenyu and Dai, Weicheng and Min, Yifei and Liu, Fenglin and Clifton, David and Zhou, S. Kevin and Staib, Lawrence and Duncan, James},
 booktitle = {Advances in Neural Information Processing Systems},
 editor = {A. Oh and T. Naumann and A. Globerson and K. Saenko and M. Hardt and S. Levine},
 pages = {9984--10021},
 publisher = {Curran Associates, Inc.},
 title = {Rethinking Semi-Supervised Medical Image Segmentation: A Variance-Reduction Perspective},
 volume = {36},
 year = {2023}
}

@ARTICLE{11150469,
  author={Zhao, Weiren and Zhong, Lanfeng and Liao, Xin and Liao, Wenjun and Zhang, Sichuan and Zhang, Shaoting and Wang, Guotai},
  journal={IEEE Transactions on Medical Imaging}, 
  title={MetaSSL: A General Heterogeneous Loss for Semi-Supervised Medical Image Segmentation}, 
  year={2025},
  volume={},
  number={},
  pages={1-1}}

@article{DBLP:journals/corr/GutmanCCHMMH16,
  author       = {David A. Gutman and
                  Noel C. F. Codella and
                  M. Emre Celebi and
                  Brian Helba and
                  Michael A. Marchetti and
                  Nabin K. Mishra and
                  Allan Halpern},
  title        = {Skin Lesion Analysis toward Melanoma Detection: {A} Challenge at the
                  International Symposium on Biomedical Imaging {(ISBI)} 2016, hosted
                  by the International Skin Imaging Collaboration {(ISIC)}},
  journal      = {CoRR},
  volume       = {abs/1605.01397},
  year         = {2016},
  eprinttype    = {arXiv},
  eprint       = {1605.01397},
  bibsource    = {dblp computer science bibliography, https://dblp.org}
}

@InProceedings{10.1007/978-3-030-37734-2_37,
author="Jha, Debesh
and Smedsrud, Pia H.
and Riegler, Michael A.
and Halvorsen, P{\aa}l
and de Lange, Thomas
and Johansen, Dag
and Johansen, H{\aa}vard D.",
editor="Ro, Yong Man
and Cheng, Wen-Huang
and Kim, Junmo
and Chu, Wei-Ta
and Cui, Peng
and Choi, Jung-Woo
and Hu, Min-Chun
and De Neve, Wesley",
title="Kvasir-SEG: A Segmented Polyp Dataset",
booktitle="MultiMedia Modeling",
year="2020",
publisher="Springer International Publishing",
address="Cham",
pages="451--462",
isbn="978-3-030-37734-2"
}

@inproceedings{NEURIPS2021_64f1f27b,
 author = {Xie, Enze and Wang, Wenhai and Yu, Zhiding and Anandkumar, Anima and Alvarez, Jose M. and Luo, Ping},
 booktitle = {Advances in Neural Information Processing Systems},
 editor = {M. Ranzato and A. Beygelzimer and Y. Dauphin and P.S. Liang and J. Wortman Vaughan},
 pages = {12077--12090},
 publisher = {Curran Associates, Inc.},
 title = {SegFormer: Simple and Efficient Design for Semantic Segmentation with Transformers},
 volume = {34},
 year = {2021}
}

@article{feng2022dmt,
  title={Dmt: Dynamic mutual training for semi-supervised learning},
  author={Feng, Zhengyang and Zhou, Qianyu and Gu, Qiqi and Tan, Xin and Cheng, Guangliang and Lu, Xuequan and Shi, Jianping and Ma, Lizhuang},
  journal={Pattern Recognition},
  volume={130},
  pages={108777},
  year={2022},
  publisher={Elsevier}
}
\end{document}